\pdfoutput=1

\documentclass[11pt]{article}

\usepackage[]{acl}

\usepackage{times}
\usepackage{latexsym}
\usepackage{graphicx}
\usepackage{epstopdf}
\usepackage{algorithm}
\usepackage{booktabs}
\usepackage{multirow}
\usepackage{array}
\usepackage{subfigure}
\usepackage{algpseudocode}
\usepackage{amsmath}
\usepackage{graphics}
\usepackage{epsfig}
\usepackage{color}
\usepackage{bm}
\usepackage{amsmath}
\usepackage{multirow}
\usepackage{graphicx}
\usepackage{array}
\usepackage{subfigure}
\usepackage{algpseudocode}
\usepackage{amsmath}
\usepackage{graphics}
\usepackage{epsfig}
\usepackage{booktabs}

\usepackage{color}
\usepackage{bm}
\usepackage{amsfonts}
\usepackage{amsmath}
\usepackage{hyperref}

\usepackage{algorithm,algpseudocode}
\usepackage[T1]{fontenc}
\usepackage{hyperref}

\usepackage[utf8]{inputenc}

\usepackage{microtype}
\usepackage{hyperref}

\title{LingYi: Medical Conversational Question Answering \\  System based on  Multi-modal Knowledge Graphs}

\author{
	Fei Xia$^{1,2}$\thanks{* These authors contribute this work equally.}, Bin Li$^{3*}$, Yixuan Weng$^{1*}$, Shizhu He$^{1,2}$, \\\textbf{Kang Liu}$^{1,2}$, \textbf{Bin Sun}$^{3}$, \textbf{Shutao Li}$^{3}$, \textbf{Jun Zhao}$^{1,2}$ \\ [0.1cm]
	$^{1}$ National Laboratory of Pattern Recognition, Institute of Automation, CAS  \\
	$^{2}$ School of Artificial Intelligence, University of Chinese Academy of Sciences  \\
	$^{3}$ College of Electrical and Information Engineering, Hunan University \\
	{xiafei2020@ia.ac.cn}, {wengsyx@gmail.com}, \\
	{\{libincn, shutao\_li, sunbin611\}@hnu.edu.cn},
	{\{shizhu.he, kliu, jzhao\}@nlpr.ia.ac.cn}
}

\begin{document}
\maketitle
\begin{abstract}
The medical conversational system can relieve the burden of doctors and improve the efficiency of healthcare, especially during the pandemic. This paper presents a medical conversational question answering (CQA) system based on the multi-modal knowledge graph, namely \textbf{``LingYi''}, which is designed as a pipeline framework to maintain high flexibility. Our system utilizes automated medical procedures including medical triage, consultation, image-text drug recommendation and record. To conduct knowledge-grounded dialogues with patients, we first construct a \textbf{C}hinese \textbf{M}edical \textbf{M}ulti-\textbf{M}odal 
\textbf{K}nowledge \textbf{G}raph {(CM3KG)} and collect a large-scale \textbf{C}hinese \textbf{M}edical \textbf{CQA} {(CMCQA)} dataset. Compared with the other existing medical question answering systems, our system adopts several state-of-the-art technologies including medical entity disambiguation and medical dialogue generation, which is more friendly to provide medical services to patients. In addition, we have open-sourced our codes which contain back-end models and front-end web pages\footnote{\url{https://github.com/WENGSYX/LingYi}}. The datasets including CM3KG\footnote{\url{https://github.com/WENGSYX/CM3KG}} and CMCQA\footnote{\url{https://github.com/WENGSYX/CMCQA}} are also released to further promote future research.

\end{abstract}
\section{Introduction}
Conversation question answering (CQA) system is an emerging research topic, it is the natural evolution of the traditional question answering (QA) paradigm  \cite{INR-074,ghazarian-etal-2021-discol}, allowing more natural conversational interactions between patients and the systems \cite{Zaib2021ConversationalQA}. CQA can improve the patients' experience by present conversational interaction \cite{zhou-etal-2021-crslab}. It can be applied to many scenarios such as electricity business \cite{Meng2021ResearchOS}, medical healthcare     \cite{Liu2021HeterogeneousGR}, and personal assistants \cite{Uurlu2020ASV}, etc. 
\par
\begin{figure}[t]
	\centering
	\includegraphics[scale=0.45]{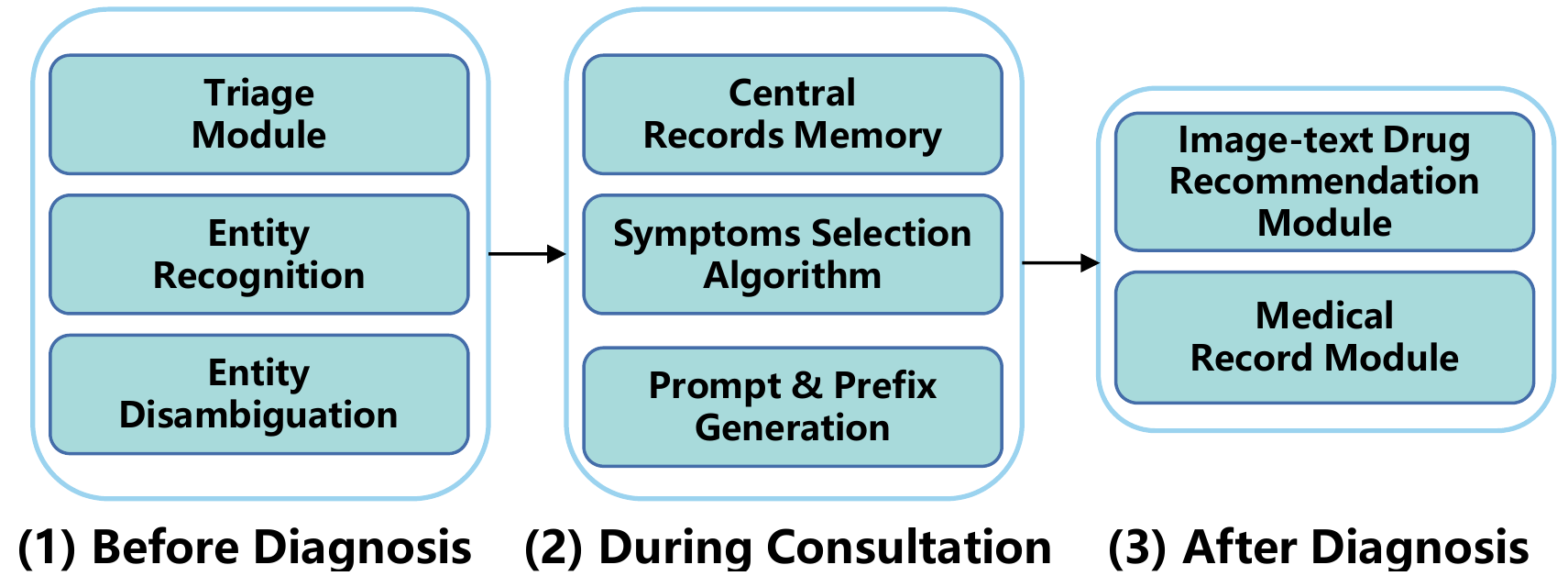}

	\caption{Main processes of the LingYi system.}
	\label{fig1}
	\vspace{-0.4cm}
\end{figure}

\begin{figure*}[t]
	\centering
	\includegraphics[scale=0.40]{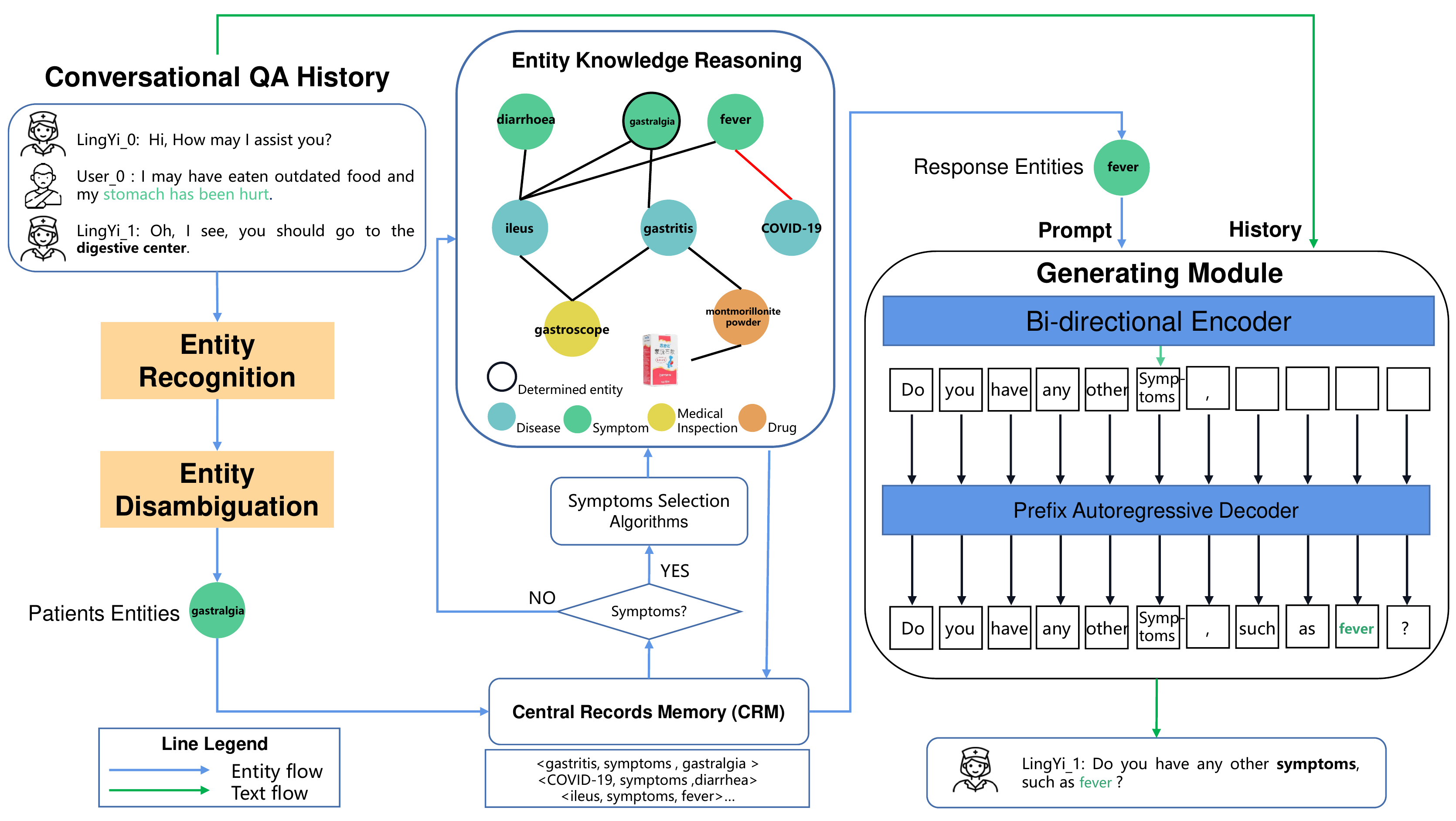}
	\caption{Overview of the medical conversational QA system of LingYi.}
	\label{overview}
	\vspace{-0.4cm}
\end{figure*}
With the pandemic of the COVID-19, it is significant for building the medical CQA system, which is advantageous to improving the efficiency of medical services and reducing the burden on doctors with broad application prospects \cite{Palanica2019PhysiciansPO}.  Recently, related medical service applications have become more and more popular, such as automatic diagnosis \cite{wang-etal-2021-covid,MOREIRA201923}, Identify symptoms \cite{ZHENG2017120}, medical image recognition \cite{C2021IdentificationOM}, etc. Most of these applications are equipped with external medical KBs for QA services, which are required to respond with manual-designed text templates. Moreover, the labor of constructing the templates is huge, where the fixed-type text responses sometimes lead to the bad patients experience. How to design a CQA system with external knowledge remains a great challenge.
\par
In this paper, we present a medical conversational question answering system with the multi-modal knowledge graph, namely LingYi, which is designed in a pipeline manner for high flexibility. As shown in Figure \ref{fig1}, it presents three main processes in our system: before diagnosis, during Consultation and after diagnosis. The before diagnosis phase consists of the triage module, entity recognition and disambiguation. The consultation phase includes Central Records Memory, Symptoms Selection Algorithm and prompt\&prefix generation. The after diagnosis phase focuses on image-text recommendation and medical record summary. In summary, LingYi provides patients with a natural cnversational QA services, which is relatively rare \cite{wang-etal-2021-covid} but more meaningful \cite{LIU202042}.

\par LingYi has the following highlights:
\begin{enumerate}
	\vspace{-0.1cm}
	\item LingYi implements a pipelined manner for automating medical procedures, providing patients a more friendly and helpful medical service with the natural conversational QA.\vspace{-0.1cm}
	\item Multiple functions such as medical triage, consultation, image-text drug recommendation and record are integrated into LingYi. In addition, our framework combines the first Chinese medical multi-modal knowledge graph CM3KG and large-scale medical conversation datasets CMCQA\footnote{\url{https://github.com/WENGSYX/LingYi}}.
	\vspace{-0.2cm}
	\item LingYi integrates advanced technologies such as entity disambiguation and medical response generation. It is competitive with other state-of-the-art (SOTA) on both automated and manual evaluations.
\end{enumerate}
\par
	\vspace{-0.2cm}
LingYi aims at providing automated medical services for the majority of patients. Preliminary experiments have been performed in the Xiangya Hospital of Central South University (Changsha, China), which demonstrates the research prospects and practical applications of the proposed system.

\section{System Description}
The Figure \ref{overview} overview the main framework of our proposed system, and the whole process include: 1) LingYi's input is the conversational QA history, and the patients' entities are obtained through entity recognition and entity disambiguation. 2) The patients' entities are then sent to the central memory recording (CRM) module for storage. 3) The dynamic symptom selection algorithm and entity knowledge reasoning are used to obtain the inference results and update the CRM. 4) The reasoned entities from the CRM and the conversation QA history are combined by prompt \cite{jin2022instanceaware} for response generation. 5) Finally, the response is generated through the bi-direction encoder and the prefix \cite{li-liang-2021-prefix} autoregressive decoder.

\subsection{Entity Disambiguation Modules}
\begin{figure}[h]
	\centering
	\includegraphics[scale=0.31]{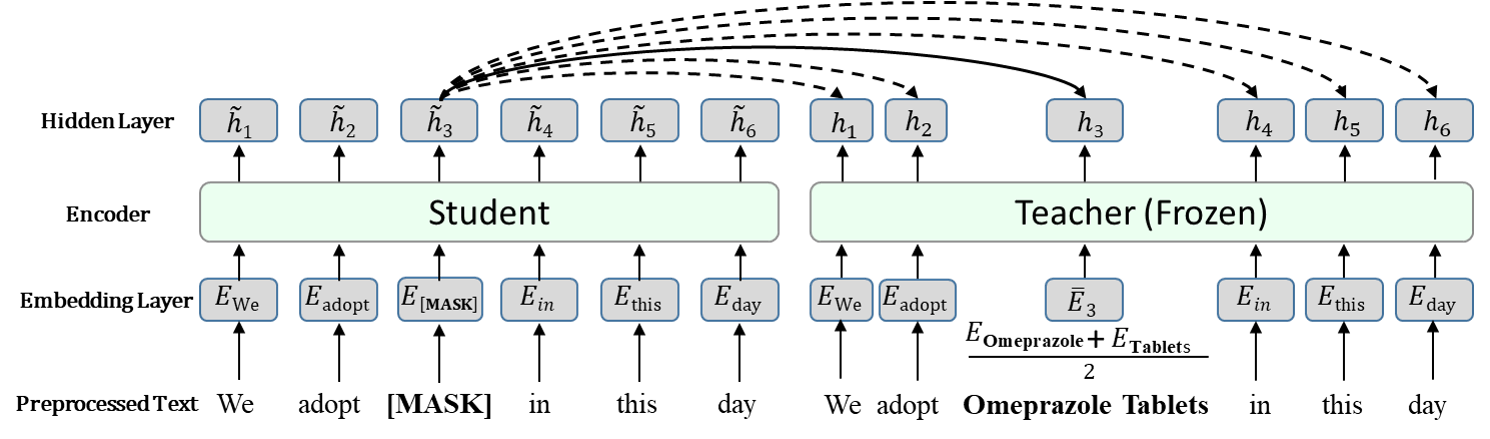}
	\caption{Contrastive pre-training in medical entity disambiguation.}
	\label{CL}
	\vspace{-0.2cm}
\end{figure}

In this section, we introduce the entity disambiguation module, which is consisted of technologies of the named entity recognition and contrastive pre-training. As for the named entity recognition, we implement the method \cite{sarker2019interpretable} for recognizing the medical entities in the utterance. We achieve the accuracy of 90.9\% in the simple medical entity recognition dataset of the IFLYTK\footnote{\url{http://challenge.xfyun.cn/topic/info?type=medical-entity}}, which ranks top-3 of the competition. After obtaining the entities, the entity disambiguation with contrastive learning is performed, which is shown in Figure \ref{CL}. We introduce the contrastive pre-training framework with Smedbert \cite{zhang-etal-2021-smedbert} for medical entity disambiguation, which is our champion scheme in SDU@AAAI-22-Shared Task2 Acronym Disambiguation\footnote{\url{https://sites.google.com/view/s-du-aaai22/home}}. Specifically, we design a contrastive pre-training method that enhances the model's generalization ability by learning the medical phrase-level contrastive distributions between true meaning and ambiguous phrases \cite{li2021simclad,DBLP:journals/corr/abs-2112-08991}. During the pre-training, we cover up the student model's medical entities, then make the student model output closer to the meaning of the teacher model, and away from other unrelated medical entities. Both two models initialize the same parameters, where the parameters of the teacher model is frozen. For the masking of these medical entities, we adopt the expert medical dictionary THUOCL \cite{han2016thuocl} for experiments. After entities are obtained, we adopt the pre-trained model for matching the recognized entities and medical entities in the knowledge graph. Finally, we map the ambiguitive phrases into the entities in the knowledge graph.

\par
\subsection{Central Records Memory}
Central Record Memory (CRM) has storage and reasoning functions, and it is mainly composed of data formats in the form of a dictionary of entity triples. First, the CRM maps the medical entities obtained in the disambiguation module to specific attributes on the knowledge graph, and stores the past entites into the dictionary. In the next round of dialogue conversation, CRM will not only map the entities of the current round on the graph but also update the current state. The information of the entities on the current round are appended into the new dictionary again. After that, the CRM will send the entities obtained by the knowledge graph inference into the generation module for further sentence generation.

\begin{algorithm} [t]
	\caption{Dymamic Symptom Selection} 
	\begin{algorithmic}
		\Require A known patient symptoms $KS$; A chinese medical multi-modal knowledge graph $KG$, The attributes in $KG$ contain $KG.symptoms$ and $KG.diseases$.
		\Ensure  \textit{Symptoms} of this round of inquiry 
		
		\State SD $\gets$ \{\}
		
		\For{$disease \in KG.disease$ }
		
		\If{$disease$ $\cap$ $KS$ = $\emptyset$}
		\For{$ds$ $\in$ disease.symptoms}
		\State SD[$ds$] \textbf{append} $disease$
		\EndFor
		\EndIf
		\EndFor
		\For{$s \in SD.keys()$}
		\If{len(SD[$s$]) >{ $\forall $}len(SD.values())}
		\State \textbf{break}
		\EndIf
		
		\EndFor
		\State \textit{Symptoms} $\gets$ $s$ \\
		\Return \textit{Symptoms}
	\end{algorithmic} 
	
	\label{algorithm1}
\end{algorithm}

\begin{table*}[t]
	\centering 
	\renewcommand\arraystretch{1.1}
	\setlength{\tabcolsep}{3.1mm}
	
	\begin{tabular}{c|cccc}
		\noalign{\hrule height 1pt}
		\textbf{Dataset} & \textbf{Domain}  & \textbf{Entity} & \textbf{Symptom} & \textbf{Dialogue} \\ 
		\noalign{\hrule height 0.5pt}
		COVID-19-CN \cite{yang2020generation} &   COVID-19       & / & / &/\\
		MedDG \cite{liu2020meddg} &  Gastroenterology      & 160  & 12  & 17864 \\
		Chunyu \cite{lin2020graphevolving} & /       & 5682  & 15 & 12842 \\
		MedDialog-CN \cite{zeng-etal-2020-meddialog} &29 Departments        & /  & 172 & \textbf{3407494} \\
		M$^{2}$ MedDialog \cite{wang2021pre} &40 Departments         & 4728  & 843 & 95408 \\
		\noalign{\hrule height 0.5pt}
		\textbf{CMCQA(Ours)} & \textbf{45 Departments}  & \textbf{33615}   & \textbf{8808} & 1294753 \\
		\noalign{\hrule height 1pt}
	\end{tabular}
	\caption{Statistics of the CM3KG compared with other datasets.}
	\label{table11}
	\vspace{-0.2cm}
\end{table*}
\begin{table}[t]
	\centering
	\renewcommand\arraystretch{1.8}
	\setlength{\tabcolsep}{0.9mm}
	
	\begin{tabular}{c|ccccc}
		\noalign{\hrule height 1pt}
		\textbf{Name} & \textbf{Symptom} & \textbf{Check} & \textbf{Drug} & \textbf{Food} & \textbf{Img} \\ 
		\noalign{\hrule height 0.5pt}
		\textbf{Num}      & 8808  & 3353        & 17318  & 366 & 3770       
		\\
		\noalign{\hrule height 1pt}
	\end{tabular}
	\caption{Statistics of entities in CM3KG}
	\label{table1}
	\vspace{-0.4cm}
\end{table}
	\vspace{-0.2cm}

\subsection{Symptoms Selection Algorithm}
In order to reduce the redundant asking rounds as much as possible and ensure accurate diagnosis of the disease, we designed a symptom selection algorithm based on dynamic programming to solve the optimization problem without recursively solving all sub-problems in turn and avoiding unnecessary calculations. As shown in the Algorithm \ref{algorithm1}, we regard each round of consultation as a sub-question to be judged, that is, we only need to select the symptom in the current state that can rule out the most diseases at one time. We traverse all the diseases in the knowledge graph, and if the intersection of the symptoms of the disease and the symptoms of the patient is not an empty set, it will be added to the list of suspected diseases. Once the symptoms of all suspected diseases are counted, the symptom with the most frequent occurrences will be found and the symptom can be judged as the output symptom. When disease reasoning is required, the CRM will perform this algorithm until the final state of the patient's disease is confirmed.

 \subsection{Entity Knowledge Reasoning}\label{Entity Knowledge Reasoning}
As for the entity knowledge reasoning, LingYi will strictly abide by the actual consultation process \cite{ha2010doctor}. The first stage is symptom reasoning, the second stage is examination reasoning, and the third stage is drug reasoning. Our system will initially conduct repeated symptom consultation to patients to ensure that the system sends complete patient symptoms entities to the CRM.  After this, LingYi will synthesize all symptoms entities from the CRM, reasoning on the basis of related entities in our CM3KG to get the patient’s medical examination. Finally, if the patient continues to consult with drug for the treatment of the disease, LingYi will preform drug recommendations with corresponding image based on CM3KG, so that the patient can find the right suggestions. In order to avoid misdiagnosis, in the symptom reasoning stage, the LingYi system will ask the patient about the symptoms in a ``diagnosed'' style until the patient's disease is confirmed.

\subsection{Generating Modules}
We adopt the method of entity prompt learning for training and prediction \cite{liu2021pre}. More precisely, we append the reasoned entities input with the conversational QA history, forming a prompt for response generation. Moreover, we design the prefix template for auto-regressive decoding. Specifically, we manually design templates of different reasoning processes described in the section \ref{Entity Knowledge Reasoning} to further increase the controllability of the generated responses. In this way, we use the prompt and prefix method to fuse the context information with the reasoned entities from CRM. As a result, the generated response will be the condition on the prompt and prefix, so as to improve the factual accuracy and controllability of the model.
\begin{table}[t]
	\centering
	\renewcommand\arraystretch{1.2}	\setlength{\tabcolsep}{1.2mm}
	
	\begin{tabular}{cccc}
		\noalign{\hrule height 1pt}
		Method     & Pre.  & Rec.  & \ \ F1 \\ \noalign{\hrule height 0.5pt}
		RoBERTa \cite{liu2019roberta}  & 0.83       & 0.74   &\ \  0.78   \\ 
		hdBERT \cite{zhong2021leveraging}   & 0.88       & 0.85    &\ \ 0.86   \\
		BERT-MT \cite{pan2021bert}    & 0.90       & 0.87   &\ \ 0.89   \\
		\noalign{\hrule height 0.5pt}
		Ours  & \textbf{0.92}      & \textbf{0.90}    & \ \ \textbf{0.91}   \\
		\noalign{\hrule height 1pt}
	\end{tabular}
	\caption{F1 performance in entity disambiguation.}
\vspace{-0.4cm}
	\label{sci}
\end{table}
\begin{table*}[t]
	\centering \small
	\renewcommand\arraystretch{1.2}
	\setlength{\tabcolsep}{1.3mm}
	\begin{tabular}{lp{1.2cm}cccm{1.2cm}cccc}
		\noalign{\hrule height 1pt}
		\vspace{-0.1cm} 
		& \multicolumn{4}{c}{CCKS A}& \multicolumn{4}{c}{CCKS B}  \\
		\cmidrule(lr){2-5} \cmidrule(lr){6-9}
		Model                 & \hspace*{0.2em}Avg. & F1 & BLEU &Dist. & \hspace*{-0.1em} Avg. & F1 & BLEU &Dist.\\ \hline
		GPT2-Entity  \cite{liu2020meddg}       &     13.43       &  25.75  &    7.30   &   7.23    &   12.41  &  24.41      &   5.81 &  7.01 \\
		HERD-Entity  \cite{liu2020meddg}       &      13.85    &  26.42  &    7.37    &   7.75   &   13.11          & 25.11  &  6.61     &   7.61    \\
		BertGPT-Entity   \cite{lewis2019bart}            &      13.79     & 26.57   &    7.03    &   7.78 &     13.69
		& 26.74   &  6.66  &   7.69       \\
		CPM2-prompt \cite{cpm-v2}  &  15.21       &  26.38  &    10.04    &   9.21   &     15.76        & 27.10   &  10.78    &   9.41    \\
		Ours     &      \textbf{17.73}       &   \textbf{30.24}  &     \textbf{12.55}    &    \textbf{10.42}    &     \textbf{18.21}     &  \textbf{30.59}    &  \textbf{13.13}   &    \textbf{10.91}       \\
		\noalign{\hrule height 1pt}
	\end{tabular}
	\caption{Performance of different methods in both CCKS-A and CCKS-B test sets.}
	\label{Performance2}
		\vspace{-0.1cm}
\end{table*}

\begin{table*}[t!]
	\centering \small
	\setlength{\tabcolsep}{2.1mm}
	\begin{tabular}{lccc}
		\noalign{\hrule height 1pt}
		Model                & \multicolumn{1}{l}{Sentence Fluency} & Knowledge Correctness & \multicolumn{1}{l}{Entire quality} \\
		\hline
		GPT2-Entity \cite{liu2020meddg}          & 3.22                                  & 3.12               & 3.17                               \\		
		HERD-Entity \cite{liu2020meddg}          & 3.83                                  & 3.77               & 3.74                               \\		
		BertGPT-Entity \cite{lewis2019bart}        & 3.71                                  & 3.78               & 3.82                               \\		
		CPM2-prompt   \cite{cpm-v2}         & 4.10                                  & 4.17               & 4.15                               \\	
		Ours          & \textbf{4.14}                                 & \textbf{4.20}               & \textbf{4.19}                               \\	
		Golden Response      & 4.77                                  & 4.83               & 4.81                               \\
		\hline
		$\kappa$                    & 0.54                                  & 0.57               & 0.58      		\\
		\noalign{\hrule height 1pt}     
	\end{tabular}
	\caption{Results of human evaluation, where $\kappa$
		is the average pairwise Cohen's kappa score between annotators.}
	\label{human}
	
		\vspace{-0.5cm}
\end{table*}
\subsection{Other Function Modules}
Our LingYi system covers other functional modules, including triage, image-text drug recommendation and medical record modules, where these modules will be introduced in turn.
\subsubsection{Triage Module}
We implement the triage function with the Smedbert \cite{zhang-etal-2021-smedbert} model, which is fine-tuned in medical entity triage data provided by IFlytek\footnote{\url{http://challenge.xfyun.cn/topic/info?type=disease-claims}}. The final results can achieve the F1 values of 90.37\% for medical triage classification. Finally, we apply this method to our medical triage module.
\subsubsection{Image-text Drug Reommendation}
We utilize the CM3KG dataset and implement medical knowledge entity reasoning, linking drug entities to corresponding images to achieve image-text drug recommendations. It will be helpful for patients to find drug information more easily. 
\subsubsection{Medical Record Module}
The last module is the medical record. In order to make it easier for patients to conduct secondary treatment more conveniently and quickly, LingYi will write a medical record from the whole conversation after patients finish consultation. Specifically, we process the unstructured conversations history information based on CPT \cite{shao2021cpt} model to generate the key summarization of the patient's condition from this consultation. At the same time, this module will also process structured information stored in central records memory, such as department, examinations, drugs, and other information. Finally, two kinds of information are integrated through post-processing splicing to generate the patient's medical record.
\vspace{-0.2cm}

\par
\section{Experimental Details}
\subsection{Data Description}
\noindent \textbf{\textit{CMCQA}}\footnote{\url{https://github.com/WENGSYX/CMCQA}} is a huge conversational question-and-answer data set for the Chinese medical field, where the statistics of medical conversation datasets is shown in Table \ref{table11}. It is collected from the Chinese medical conversational question answering website ChunYu\footnote{\url{https://www.chunyuyisheng.com/}}, and has medical conversational materials in 45 departments, such as andrology, stormotologry, gynaecology and obstetrics. Specifically, CMCQA has 1.3 million complete sessions or 19.83 million statements or 0.65 billion tokens. At the same time, we further open source all data to promote the development of related fields of conversational question answering in the medical field.
\par \noindent \textbf{\textit{CM3KG}}\footnote{\url{https://github.com/WENGSYX/CM3KG}} is open-sourced multi-modal knowledge graphs. We have processed the data crawled from the website, and then sorted it into the form of tables. For example, for the symptom of stomachache, the ``disease'' attributes include ``gastritis'', ``gastric cancer'', ``gastric ulcer'' and other diseases. The ``examination'' attributes include ``gastroscopy'' and ``pathological biopsy of gastric mucosa'' , etc. After that, we search and link the entities in the knowledge graphs in Bing image database\footnote{\url{https://Bing.com/image}}. After the completion of the construction, the authors manually correct it again, eliminated about 20\% of the obvious error information, and then submit it to the expert doctors for final verification to ensure the accuracy of the multimodal knowledge graphs.
\vspace{-0.2cm}
\subsection{Implementation}
We train the model based on the Pytorch \cite{NEURIPS2019_bdbca288} and use the hugging-face \cite{wolf-etal-2020-transformers} framework. All the finetuned models are implemented in the collected medical corpus\footnote{\url{https://github.com/Lireanstar/Medical-Dialogue-Corpus}}. During training, we employ
the AdamW optimizer \cite{Loshchilov2017FixingWD}. The learning rate is set to 1e-5 with the warm-up \cite{7780459}. Four 3090 GPUs are implemented for all experiments.

\begin{figure*}[!t]
	\centering
\includegraphics[scale=0.46]{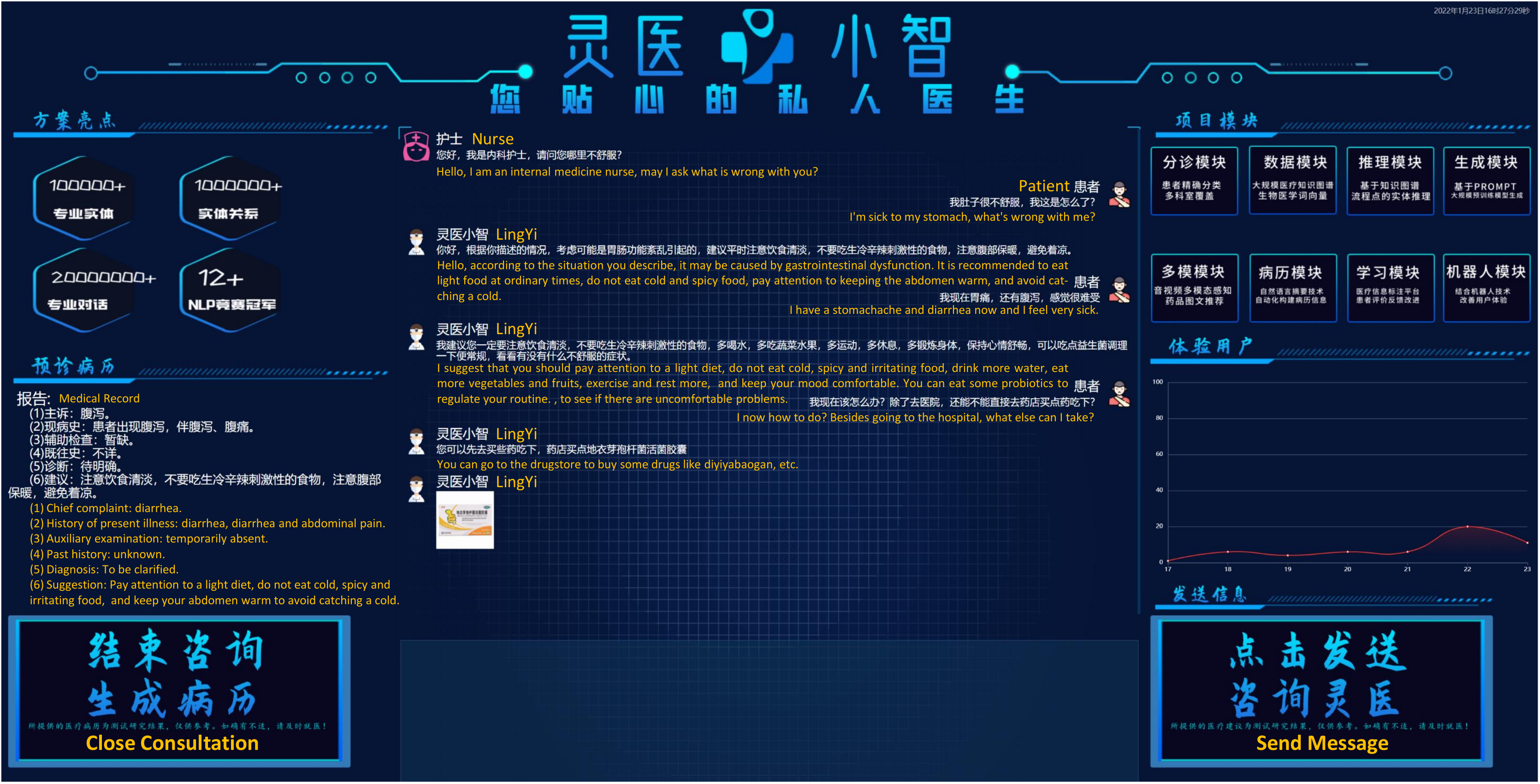}
	\caption{Snapshot of the proposed LingYi system.}
	\label{app}
	\vspace{-0.6cm}
\end{figure*}
\vspace{-0.1cm}
\subsection{Evaluation Introduction}
To ensure correct medical entity information and fluent responses, we provide the automatic and manual evaluations. Specifically, we conduct experiments entity disambiguation dataset of SDU@AAAI 2021\footnote{\url{https://sites.google.com/view/sdu-aaai22/home}} and medical dialogue generation dataset of the CCKS\footnote{\url{https://www.biendata.xyz/competition/ccks_2021_mdg/}}. We adopt the evaluation metrics, including the F1, BLEU \cite{papineni-etal-2002-bleu} , and Dist. scores. The F1 score reflects the correctness of medical entity knowledge. The BLEU score reflects the relativity of the generated responses. The Dist. score represents the diversity of the generated sentences. We further prepare manual evaluation for randomly picking 100 cases from the test dataset. Each generated sentence is scored by three independent persons with a medical background. We adopt the same human evaluation metrics as the work \cite{liu2020meddg}. The rating scale for each metric is ranged from 1 to 5, where 1 represents the worst and 5 the best.
\vspace{-0.1cm}
\par 
\subsection{Results}
The experimental results are shown in Table \ref{sci}. It can be seen that our results achieve the best results compared to other entity disambiguation SOTA methods. 
At the same time, we also conducted related evaluations on the test dataset, which is shown in Table \ref{Performance2}. As is shown from the table, our method achieves the best results against recent strong baselines and leads in accuracy, relevance and diversity. We provide a manual evaluation to further judge the performance between different methods. 
As shown in Table \ref{human}, our method achieves competitiveness in human evaluation compared to other SOTA methods. It is noted that there is still a long way from the generated responses to the real responses of people.
What's more, the average pairwise Cohen's kappa \cite{randolph2005free} scores between annotators range between 0.4 and 0.6 for all metrics, which represents a moderate annotator agreement.

\section{Application}
We present the application of LingYi at the website\footnote{\url{http://kg.wengsyx.com/} for Chinese version, \url{http://kg.wengsyx.com/lyxz_en/} for English version.}, where the snapshot are shown in Figure \ref{app}. Figure \ref{app} shows that if the patient says that he is sick in his stomach, the system will get the entity ``gassralgia'' from the entity disambiguation module. Afterwards it will obtain the entity ``gastritis'' from the knowledge graphs through entity knowledge reasoning. The reasoned entity is sent to the generating module for further recommending the patient to do diagnosis in the hospital. Finally, if a patient needs urgent drug, the system will recommend the proper drug through the knowledge graphs. A medical record will be generated after the consultation, which will significantly facilitate the patient's secondary treatment\footnote{Medical conversational QA demo: \url{https://www.youtube.com/watch?v=fsFnbim5hWc}}.

\section{Conclusion}

In this paper, we introduce a conversational medical QA system, LingYi, which integrates multi-modal knowledge graphs CM3KG, and medical conversation dataset CMCQA. We design a pipeline to analyze patients' statements and present SOTA performance on both entity disambiguation and response generation tasks, providing patients with a full range of medical consulting services. We hope our system will alleviate the problem of worldwide medical resource scarcity during the COVID-19 and provide a feasible direction for subsequent researchers to develop medical artificial intelligence.

\bibliography{anthology,custom2}
\bibliographystyle{acl_natbib}

\appendix
\section{Ethical Considerations}
	\vspace{-0.1cm}
The constructed system aims to generate professional, fluent, and consistent medical responses. We have also realized that, due to adapt of pre-trained models which learning with the medical data from Internet, the proposed approach may produce inappropriate text such as offensive, racially or gender-sensitive responses. Meanwhile, although the proposed method can cover the stages of before, during, and after the medical treatment, it may also be maliciously exploited, for example, using forged medical reports to fabricate false medical reports.
\par
  We have carefully considered the above issues and provided the following detailed explanations: (1) All used medical data is collected from the Internet, and it is inevitable contain offensive, racially or gender-sensitive doctor-patient conversations. Due to the limited space, we briefly describe the characteristics and cleaning rules of the datasets and delete the utterances of doctor-patient dialogue that are offensive, racially, or gender-sensitive. The detailed process can be found in the README file on the website \url{https://github.com/Wengsyx/LingYi}. (2) The quality of the processed datasets will affect the credibility of the robustness evaluation. Compared with previous works, we adapt four types of criteria to evaluate the credibility of our system, they are: offline index evaluation (BLEU, Distinct and F1), online patients evaluation, dialogue rounds testing, and professional doctor evaluation. We hope to maximize the reliability and implement ability of the system based on such evaluation benchmarks. (3) LingYi is a medical system with a suggestion nature, which uses knowledge graph to provide multi-modal medical responses. Our system may produce incorrect medical results. Therefore, the responses of the system are only for reference. Normal patients should not seek medical treatment indiscriminately. (4) Our work does not contain identity information, the doctor only responds by the patient's condition, it will not harm anyone and doesn't invade people’s privacy. (5) The medicines recommended by LingYi are over-the-counter medicines. Patients need to consult their doctor for further confirmation when purchasing the drugs for prescription drugs. (6) Our system supports applications on different terminals. 

In the future, we will adopt federated learning to capture the patient's condition and provide comprehensive protection more accurately, such to the federated learning is able to provide privatized and personalized learning services for each patient. Finally, since the proposed method uses external knowledge graphs, the information sources of these knowledge graphs also suffer from several issues such as risk and bias. Reducing these potential risks requires ongoing research.

\end{document}